\documentclass{article}

\usepackage{PRIMEarxiv}

\usepackage[utf8]{inputenc} 
\usepackage[T1]{fontenc}    
\usepackage{hyperref}       
\usepackage{url}            
\usepackage{booktabs}       
\usepackage{amsfonts}       
\usepackage{nicefrac}       
\usepackage{microtype}      
\usepackage{lipsum}
\usepackage{fancyhdr}       
\usepackage{graphicx}       
\graphicspath{{media/}}     

\usepackage{authblk}
\usepackage{needspace}

\title{\fontsize{15pt}{16pt}\selectfont From Data-Driven to Purpose-Driven Artificial Intelligence: Systems Thinking for Data-Analytic Automation of Patient Care
}

\author[1,2,3,*]{Daniel Anadria}
\author[2]{Roel Dobbe}
\author[1]{Anastasia Giachanou}
\author[3]{Ruurd Kuiper}
\author[3]{Richard Bartels}
\author[3]{Wouter van Amsterdam}
\author[2]{Íñigo Martínez de Rituerto de Troya}
\author[4]{Carmen Zürcher}
\author[1,3]{Daniel Oberski}

\affil[1]{Utrecht University, the Netherlands}
\affil[2]{Delft University of Technology, the Netherlands}
\affil[3]{University Medical Center Utrecht, the Netherlands}
\affil[4]{Organisation for Economic Co-operation and Development, France}
\affil[*]{First and corresponding author (\texttt{d.anadria@uu.nl}).}

\begin{document}
\maketitle

\begin{abstract}
In this work, we reflect on the data-driven modeling paradigm that is gaining ground in AI-driven automation of patient care. We argue that the repurposing of existing real-world patient datasets for machine learning may not always represent an optimal approach to model development as it could lead to undesirable outcomes in patient care. We reflect on the history of data analysis to explain how the data-driven paradigm rose to popularity, and we envision ways in which systems thinking and clinical domain theory could complement the existing model development approaches in reaching human-centric outcomes. We call for a purpose-driven machine learning paradigm that is grounded in clinical theory and the sociotechnical realities of real-world operational contexts. We argue that understanding the utility of existing patient datasets requires looking in two directions: upstream towards the data generation, and downstream towards the automation objectives. This purpose-driven perspective to AI system development opens up new methodological opportunities and holds promise for AI automation of patient care.
\end{abstract}


\section{Introduction}
\label{sec:introduction}

Machine learning is rapidly becoming the dominant modeling paradigm in biomedicine where much of the contemporary discourse on AI automation is data-driven and theory-agnostic. These two properties were once the selling points of machine learning set to supersede theory-driven modeling approaches \cite{breiman_Statistical_2001}. Over the last two decades, the role of clinical domain theory, the design of model inputs, and requirements engineering have all diminished in data-analytic model development. While the data-driven nature of machine learning has enabled fascinating predictive pipelines \cite{kline_Multimodal_2022} and unforeseen emergent properties in foundational models \cite{bommasani_Opportunities_2021}, it remains unclear when such AI systems can be considered safe for the automation of patient care \cite{obermeyer_Dissecting_2019, kucharavy_Fundamental_2024, amsterdam_When_2025}. 

The current machine learning paradigm can be understood as valuing data quantity over data quality \cite{burden_fit_2024}. While real-world patient data take on a central role in model development, developers often find it challenging to look ahead towards what tasks any particular dataset may help automate, and to look backward from an envisioned AI system functionalities to its data requirements. Both of these challenges require a grounding of model development in the broader sociotechnical context. Machine learning for AI automation requires an understanding of both clinical reasoning and of operational contexts where the AI system is intended to be used. We view both of these challenges as calling for purposeful AI system design.

Our perspective applies systems thinking onto data-analytic model development. We posit that data, models, and people in any operational setting exert a strong influence on each other as they shape systemic outcomes over time. This means that responsible machine learning has to account for some of the sociotechnical complexity and provide mechanisms that can ensure the scientific validity and temporal robustness of the proposed model solutions. This is by no means an easy task. Some complexity has to be kept contained for AI automation to proceed, but other parts cannot be abstracted away without derailing the automation from its purpose \cite{selbst_Fairness_2019}. In sum, there are complexities that need to be wrestled with, and it takes wisdom to know which ones those are.

We ground our work in patient care where large quantities of real-world patient data are increasingly repurposed for training, fine-tuning, and deployment of AI systems \cite{europeanunion_Regulation_2025}. While healthcare professionals desire automation of various work activities \cite{fruehwirt_Better_2021}, this aim is only weakly reflected in the design of current patient data pipelines. In fact, much of the ongoing data collection in patient care is shaped by its reporting, communication, and billing needs \cite{schulz_Standards_2019}. This leads to a situation where an abundance of patient data can coexist with a lack of clarity as to what patient care tasks these data can help automate.

In machine learning, model development often starts from an exploration of patient data for their utility (\textit{problem-shopping}) rather than from the intended purpose of AI automation (\textit{problem-solving}). This amounts to preparing a dish from the ingredients that are available in the fridge, rather than carefully collecting the ingredients required by a particular recipe. While this can lead to some great dishes, such an approach is inherently constrained in what it can deliver. As a consequence, proposed AI systems may lack real-world utility as they fail to address the pressing clinical problems. At worst, data-driven approaches that are not well-thought-out could even cause patient harms \cite{obermeyer_Dissecting_2019, amsterdam_When_2025}.

However, the issue of using existing patient data is not black-or-white. Since patient data are collected at scale, they may hold the potential to enable AI automation. In this work, we identify two data-centric challenges that need addressing:
\begin{enumerate}
    \item \textbf{Knowing the data we have:} How does patient data generation in clinical care shape patient data quality? (explored in Section \ref{sec:data-we-have})
    \item \textbf{Knowing the data we want:} How do the characteristics of the AI automation task influence data quality requirements? (explored in Section \ref{sec:data-we-want})
\end{enumerate}

This work seeks to deliver a nuanced message. Our main argument is that machine learning for AI automation can benefit from systems thinking. To understand the current data-driven paradigm, we compare it to the theory-driven paradigm which used to be more common but has been losing ground over the last decades. Our message is not a call for a return to theory-driven modeling. Instead, we want to emphasize its aspects that could benefit the AI automation of patient care. This represents an expansion of the current machine learning thought and paves the way for model development methods that can be considered parts of a purpose-driven, human-centric, and responsible AI system development. Our contributions are as follows:

\begin{itemize}
    \item This is the first work to combine data analysis and systems theory for AI automation. We elaborate on how systems thinking can aid responsible machine learning in patient care.
    \item We explore the functional and quality requirements of patient data as AI automation system components. 
    \item We discuss how systems thinking can support an informed use of existing patient data, and inform the design of patient data collection and processing pipelines for future AI automation. 
\end{itemize}

The remainder of this paper is structured as follows. Section \ref{background} offers background information on data analysis (Section \ref{primer-data-analysis}) and systems theory (Section \ref{primer-systems-theory}), as they relate to AI automation. In Section \ref{sec:data-we-have}, we explain how patient care and patient data generation could be seen from a systems perspective (Section \ref{sec:data-we-have-theory}), and we describe some of the key challenges in clinical data generation as they influence the quality of current patient data for AI automation (Section \ref{sec:data-we-have-theory}). In Section \ref{sec:data-we-want}, we apply systems-analytic theory to further explore how the intended purpose of AI automation shapes its data requirements. Section \ref{sec:discussion} offers a discussion of our main findings and limitations. Section \ref{sec:conclusions} concludes the paper.

\section{Background}
\label{background}

\subsection{A Primer on Data Analysis}
\label{primer-data-analysis}

\begin{figure}[h]
\centering
\caption{Model development begins with nature that generates data. In data-analytic modeling, a sample of data, produced by a real-world data-generating process, is used develop a model. The model is a mathematical representation of a part of the joint distribution produced by the data-generating mechanism that is of interest, and can be evaluated by its fit to (1) the data, and (2) the world.}
\includegraphics[width=0.93\textwidth]{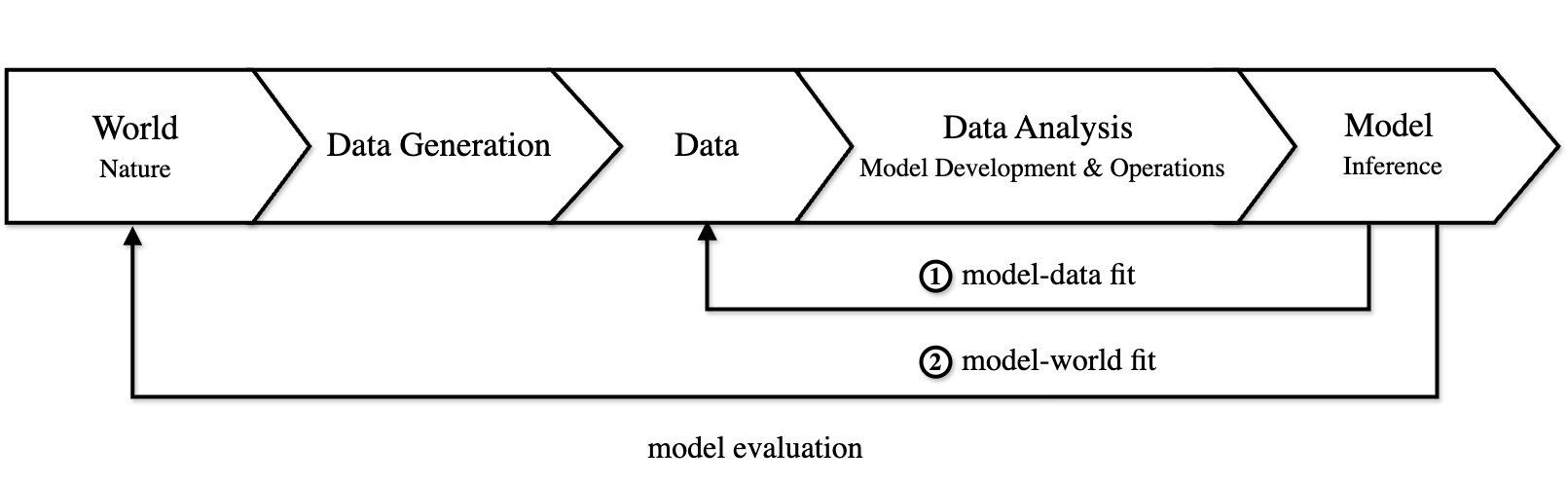}
\label{fig:statistical-inference}
\end{figure}

\subsubsection{What is data analysis?}

We use the term \textit{data analysis} and its derivatives in a broad sense to encompass all quantitative methods that are predicated on statistical inference. This field is splintered into various factions including data science, statistics, epidemiology, biostatistics, bioinformatics, psychometrics, econometrics, clinimetrics, etc. Each faction is largely shaped by its application domain. Sometimes, there is a difference in the subset of the data-analytic techniques used, and often there is a difference in naming of similar concepts. Our usage of the term data analysis is in line with how the field was first defined by Tukey \cite{tukey_Future_1962} to comprise all the quantitative techniques and the subsequently evolved factions. Data-analytic methods have become the engine of contemporary AI systems \cite{finlayson_Machine_2023}. In this work, the term data analysis is used to encompass both the initial model development and the subsequent model operations.

\subsubsection{What is statistical inference?}

Data analysis covers a wide range of techniques that leverage statistical inference on historical data. The term \textit{statistical inference}, also known as \textit{learning}, refers to ways of extracting knowledge about the world from a sample of its data \cite{geisser_Predictive_1993, cox_Principles_2006, james_Introduction_2021}. In statistical theory, inference is useful for deriving \textit{explanations} about how some components of reality relate to each other (e.g. through parameter estimation or hypothesis testing), and \textit{predictions} of the likely future states or outcomes within the modeled system \cite{james_Introduction_2021}. Statistical inference represents the logic of data-analytic model development (Figure \ref{fig:statistical-inference}). While all data-analytic models share reliance on statistical inference, model development can be approached in multiple ways. 

\subsubsection{What are the two data-analytic cultures?}

\begin{table}[h]
\centering
  \caption{Differences Between the Two Data-Analytic Cultures.}
  \label{tab:twocultures}
  \begin{tabular}{cc}
    \toprule
    \textbf{Theory-Driven Modeling} & \textbf{Data-Driven Modeling} \\
    (i.e. data modeling \cite{breiman_Statistical_2001}) & (i.e. algorithmic modeling \cite{breiman_Statistical_2001})\\
    \midrule
    Small data & Big data \\
    \midrule
    Parametric models & Nonparametric models \\
    \midrule
    Inference as a means of explanation & Inference as a means of prediction \\
    \midrule
    Emphasis on model inputs & Emphasis on model outputs \\
    \midrule
    Model development is theory-driven & Model development is theory-agnostic \\
    \midrule
    Data are carefully considered. & Data are largely taken at face value. \\
    Both data collection and model specification & Less attention is given to data collection.\\
    are informed by the sociotechnical context & Model specification prioritizes on outputs \\ 
     (e.g. via research design) & (e.g. via loss function optimization)\\
    \midrule
    Model evaluation focuses on both model- & Model evaluation largely focuses on model-data\\
    data fit (goodness-of-fit measures) and & fit (e.g. predictive performance metrics) and\\
    model-world fit (through validity of inputs) & the model-world fit is de-emphasized\\
    \midrule
     Associated with statistical modeling in & Associated with machine learning and\\
     social sciences and biomedicine & computational modeling in data science \\
    \bottomrule
  \end{tabular}
\end{table}

\needspace{2\baselineskip}
Contemporary data analysis can be characterized by two dominant model development paradigms -- the \textit{theory-driven} and \textit{data-driven modeling} \cite{breiman_Statistical_2001}. The distinction between the two model development cultures stems from the history of data analysis. The fuzzy boundary is becoming increasingly blurred with time and models need not fall neatly into these two categories. In our view, this historical distinction can be useful to capture some of the fundamental ideas affecting contemporary AI model development. Table \ref{tab:twocultures} offers a summary of the key points discussed in this segment.

Traditionally, theory-driven modeling has been the hallmark of social science research \cite{geisser_Predictive_1993, breiman_Statistical_2001}. This data-analytic culture has emerged in the contexts of data scarcity where challenges posed by small datasets had to be overcome through research design (to ensure high data quality / low error of model inputs), parametric models (to improve statistical power when analyzing small datasets), and domain knowledge (to inform what relationships ought to be modeled or studied in the first place). In this paradigm, statistical inference was largely used for explanation aims such as hypothesis testing and parameter estimation \cite{geisser_Predictive_1993}.

In contrast, data-driven modeling arose along the private sector contexts where large quantities of data were more readily available \cite{breiman_Statistical_2001}. The abundance of data has had an effect on model development where non-parametric models could infer the functional form of the learned model relationship directly from the data. The data-driven modeling culture, including much of contemporary machine learning, has de-emphasized reliance on domain theory in input design (e.g. data collection, feature engineering). Instead, this paradigm emphasizes forecasting or predictive inference and hence focuses on model outputs \cite{breiman_Statistical_2001}. 

Each culture evaluates models differently. Conceptually, models can be assessed along the \textit{model-data fit} and along the \textit{model-world fit} axes \cite{cox_Principles_2006}. Both cultures examine the model-data fit. In theory-driven models, this is may be done through goodness-of-fit measures, and in data-driven modeling, through predictive accuracy \cite{breiman_Statistical_2001}. However, these two traditions differ in their approach to the model-world fit in important ways.

In both cultures, parts model development are influenced by applied domain theory. This is especially the case in model specification where an appropriate functional form for the model representation has to be chosen. For instance, understanding that clinical imagining classification is a translation equivariant task \cite{anwar_Medical_2018} points to the possible appropriateness of using convolutional neural network architectures \cite{lenc_Understanding_2019}. 

In theory-based modeling, domain theory also has a profound effect on aspects of data generation, including data collection, sampling, and feature engineering. In this culture, the validity of the learned inferences is understood to flow from input design. Model inputs (data) are assumed to have high validity and low measurement error thanks to the rigor that goes into their preparation. 

In much of the data-driven tradition, model inputs (data) are largely taken as a given and attention is shifted further downstream in the model development. The quality of the model is largely examined through its predictive performance on a held-out validation set \cite{breiman_Statistical_2001}. From the confusion matrix, one can derive a plethora of model performance metrics such as its accuracy and AUC, as well as the algorithmic fairness metrics such as demographic parity and equal opportunity \cite{riccilara_Addressing_2022}. 

While model outputs certainly capture parts of the greater model quality, by overlooking data generation, machine learning may risk abstracting away important parts of the model-world fit. To explore this further, we turn our attention to data generation as it is understood by data analysis.

\subsubsection{What is a data generating mechanism?}

Statisticians and data scientists use the term \textit{data-generating mechanism} (or simply \textit{data generation}) to refer to a complex, elusive, and perhaps not entirely knowable process that underlies the production and explains the structure of observed data (see Figure \ref{fig:statistical-inference}). This term has a philosophical dimension that exceeds the simple notions of data collection and processing and bewilders statisticians. Data generation speaks to the very core of what data-analytic modeling attempts to do - representing natural systems of interest through mathematical equations. In computational statistics, one may have complete control over the data generating mechanism in simulated settings. For example, we could define a random variable $y$ as:
\begin{equation}
    y = 3 + 1.3x + 0.3x^2
\end{equation}

where $x$ as is Gaussian random variable with the following mean and standard deviation:

\begin{equation}
    x \sim \mathcal{N}(\mu = 100, \sigma=15)
\end{equation}

\needspace{3\baselineskip}
In this setup, we know the probability density function of $x$ is:

\begin{equation}
f(x) = \frac{1}{\sqrt{450\pi}} e^{-\frac{(x - 100)^2}{450}}
\end{equation}

These three equations give us the exact data-generating mechanism of $y$. Having this knowledge enables us to simulate a dataset by randomly sampling $x$ and computing $y$. We could add further complexity to this and create a large dataset with many related variables and some stochastic noise in the equation. As creators, we would still know the configuration of the true data-generating mechanism. However, this level of certainty about data generation is rarely possible when dealing with real-world data. 

In most applications, we simply do not know the true data-generating mechanism. In such cases, data analysis can be used to estimate this process, or to model a part of the multivariate distribution that it produces. Perhaps the best description of the data-generating mechanism for real-world data is given by Breiman \cite{breiman_Statistical_2001}:
\begin{quote}
\begin{center}
\texttt{"Think of the data as being generated by a black box in which a vector of input variables x (independent variables) go in one side, and on the other side the response variables y come out. Inside the black box, nature functions to associate the predictor variables with the response variables [\ldots]" \\ 
-- Leo Breiman, 'Statistical Modeling: The Two Cultures' (2001)}
\end{center}
\end{quote}

This quote reveals two valuable insights of the data-analytic perspective on data generation. First, the claim that real-world data generation is a black box means that it is a complex system whose internal workings are hidden or not readily understood. Second, discussed in greater detail by Breiman \cite{breiman_Statistical_2001}, is that 'to model' data (as in to \textit{develop}, \textit{fit} or \textit{train} a model) means to represent or leverage some parts of the structure that is defined by their generating mechanism. Data-analytic models are thus theories about ways in which natural processes function. 

\subsubsection{How do the two cultures approach data generation?}

If data-analytic models are theories about the world, they have to match the natural system that is being modeled in some meaningful way. The two aforementioned model development approaches, theory-driven and data-driven modeling, differ in ways in which they evaluate this fit (Figure \ref{fig:statistical-inference}). 

Many theory-driven approaches emphasize the design of \textit{model inputs} (e.g. one may exert control over data generation through experimental design). This may be familiar to readers with backgrounds in social sciences and statistics. Since datasets in this tradition tend to be smaller, models often times come with parametric assumptions that increase their statistical power. Model development frequently leans on domain theory and sociotechnical context to inform data collection and model specification. In many cases, this results in a certain level of design to model development. However, this outcome is by no means caused by small datasets nor parametric model architectures. Rather, it is an artifact of a model development culture that assigns high value to design of its inputs \cite{breiman_Statistical_2001}.

In contrast, much of data-driven modeling can be understood as prioritizing \textit{model outputs} (esp. predictive performance on the test set). This data-analytic culture may be familiar to readers with backgrounds in machine learning and data science. Data are largely taken at face value and models are trained to have high predictive performance on a held-out validation set \cite{breiman_Statistical_2001}. Historical circumstances in data analysis have led the data-driven culture to de-emphasize design approaches and evaluations of the model-world fit \cite{breiman_Statistical_2001, cox_Principles_2006}. However, this too is just a cultural artifact as there is nothing fundamental to big data nor complex model architectures that prevents them being supplemented with purposeful design. 

\subsubsection{Why AI automation requires an understanding of data generation?}

Given the increased data-driven nature of AI automation, there is a need to reevaluate common model development practices. If models ought to learn trustworthy representations of reality from data, then their evaluation ought to not rest solely on their inputs nor outputs. Model development should carry a strong understanding of the real-world data generation and sociotechnical reality of the applied context to ensure the validity of AI system design. 

Works that are sufficiently cognizant of both of the aforementioned aspects are still rare in academic machine learning literature. All too often, data alone are used to evaluate model quality. In such instances, model developers sometimes do not sufficiently reflect on the real-world processes whose parts their models might be representing, nor on how their models fit to and transform that same reality \cite{joshi_AI_2025}.

The message here is nuanced. There is nothing inherent to small datasets nor parametric models that guarantees the learning of better model representations, nor that leads to improved real-world automation outcomes. Model complexity and data quantity are not the key determining factors of these ends. One could certainly fit a parametric model without much sociotechnical awareness and domain knowledge. 

Similarly, some data-driven architectures can fit real-world problems quite well. 

In other cases, however, data-analytic models of any complexity may appear to fit the data well, yet result in harms when they fail to imitate the right components of reality in deployment \cite{obermeyer_Dissecting_2019, amsterdam_When_2025}. 

The secrets to what parts of reality the data measure, and what relationships ought to be learned and enacted by AI systems both lie hidden in the \textit{nature's black box} \cite{breiman_Statistical_2001}. While certainly useful for the study of nature, there are limits to what data analysis alone can reveal. Consider that most data-analytic models cannot differentiate between correlations and causal relationships well \cite{richens_Improving_2020}. This also means that data-analytic knowledge is rarely sufficient to differentiate between legitimate biomedical signal and undesirable data biases \cite{anadria_Algorithmic_2024}. 

For example, a data-analytic model may learn from data that family visits are positively correlated with patient mortality \cite{gianfrancesco_Potential_2018}. If the data-generating mechanism is such that nurses encourage family visits for those patients which appear to be more seriously ill, then this is a correlation. Asking the family not to come visit would not change the patient's odds of survival. It is helpful to know the context surrounding data generation. The sheer presence of algorithmic bias makes it evident that data should not speak for themselves when all that is known about data are correlations \cite{barocas_Fairness_2023}.

\subsection{A Primer on Systems Theory}
\label{primer-systems-theory}

\subsubsection{What is systems theory?}
\label{sec:what-is-systems-theory}

Systems theory is a research paradigm that tries to approach the complexity of the world through a holistic lens which emphasizes relationships of its many interconnected components. It is a versatile language and a way of reasoning that is intended to inform meaningful system change and purposeful system design \cite{anderson_Systems_1997, leveson_Engineering_2016, enserink_Policy_2022}. 

\subsubsection{What is systems thinking?}
The systems paradigm is transdisciplinary. In fact, a system can be defined freely as any set of components which perform a joint function \cite{leveson_Engineering_2016}. This flexibility has enabled \textit{systems thinking} -- the application of system-theoretic principles onto various applied domains. This has been done in physics, biology, policy research, and urban planning. However, to our knowledge, systems thinking has not been applied to data-analytic model development for AI automation prior to this work.

\subsubsection{What system-theoretic approaches do we use?}
Systems theory is vast, and in this work, we discuss only a small subset of its ideas and argue for their value in advancing data-analytic model development. We primarily engage with ideas that arose within control systems engineering \cite{leveson_Engineering_2016}, systems safety for AI \cite{dobbe_Hard_2021, dobbe_System_2024}, and systems analysis for policy research \cite{enserink_Policy_2022}.

\subsection{How can systems thinking be applied to data-analytic automation of patient care?}
\label{sec:premises}

We base our further work on the following three premises:

\begin{enumerate}
    \item Patient care is a complex system that operates with the fundamental aim of promoting patient health. This system contains clinical decision-making processes that give rise to downstream patient outcomes. 
    \item Patient data generation is a sociotechnical component of patient care that operates to collect, interpret, and manage patient data. It can be understood as a complex system in its own right whose operations shadow and often imperfectly mirror clinical decision-making and downstream patient outcomes. We primarily focus on its data collection component (as it links to data analysis), rather than factors related to disease etiology.
    \item AI automation is a complex engineering problem that seeks to augment patient care through the introduction of data-analytic models into clinical decision-making.
\end{enumerate}

The following two sections make use of these systems definitions. Section \ref{sec:data-we-have} applies a broad systems lens to explore how patient care processes shape patient data generation and the emergent data quality. Section \ref{sec:data-we-want}, focuses on the AI automation as it shapes data quality requirements. Taken together, these system-theoretic takes on data quality for AI automation can help understand the value of data we have, and identify requirements for the data we want as informed by automation objectives.
\section{Data We Have: Systems Theory for Patient Data Generation}
\label{sec:data-we-have}

\subsection{Context}
\label{sec:data-we-have-premise}

\begin{figure}[h]
    \begin{center}
        \caption{Patient data generation is a complex sociotechnical system within patient care. Through data collection, measurements of patient health states, clinical narratives, and treatment decisions all become encoded in various information artifacts which we consider raw patient data.}
        \includegraphics[width=0.65\textwidth]{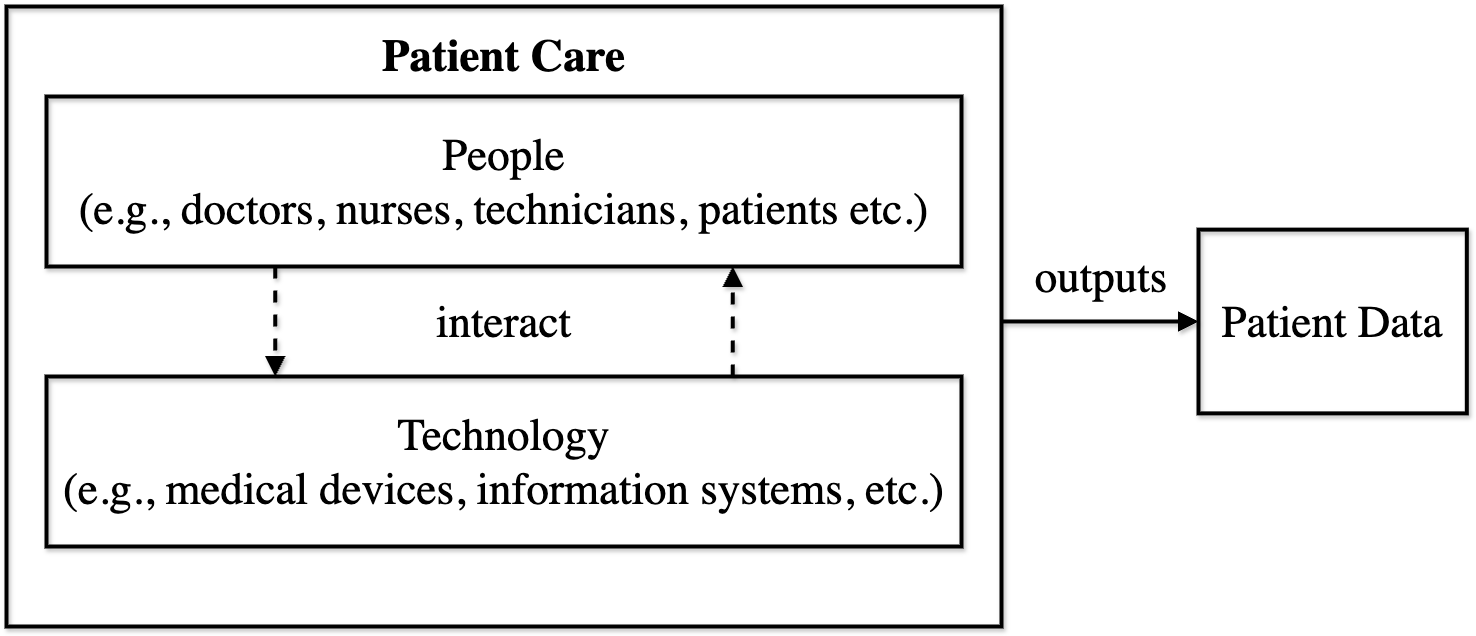}
        \label{fig:datageneration}
    \end{center}
    \end{figure}

In this section, we explore how some of the key characteristics of real-world patient data generation influence the quality of current patient data. We argue that an application of a broad systems lens onto the issue of patient data generation can help inform purpose-driven AI model development. This approach uncovers aspects of reality that strongly qualify patient data we have and which cannot be easily learned through data analysis alone.

Rather than mapping the generation of any particular patient dataset, we prioritize discussing some of the commonly shared properties among them. This amounts to looking upstream from the current data towards the processes that had produced them. Our perspective is inspired by control systems theory \cite{leveson_Engineering_2016} where sociotechnical systems can be understood in terms of controlled processes (both clinical decision making and patient data collection) and controllers (people and technology involved in patient care). 

As alluded to in Section \ref{sec:premises}, the key premise of this section is that patient care can be understood as a complex sociotechnical system that operates with the goal of promoting patient health \cite{harteloh_Quality_2003}, and that contains a data-generating subsystem (Figure \ref{fig:datageneration}). We study how the two systems relate to one another and how their interactions shape the quality of current patient data for AI automation.

\subsection{A Systems Perspective on Patient Data Generation}
\label{sec:data-we-have-theory}

Individuals become patients only when they seek clinical care. The care-seeking behavior often follows a health event that often represents a perceived negative change in the individual's well-being. However, the decision of whether to seek clinical care or not also represents a filter that influences the downstream representativeness of patient data. 

The mechanism which differentiates between those patients that seek care and those that do not is complex and varies along personal and societal characteristics. In some cases, parts of this mechanism could fall within an individual's discretion. For instance, men appear to seek less care than women in some societies \cite{galdas_MenHealthHelpseeking_2005}, and members of some religious denominations may avoid certain types of clinical care \cite{gohel_How_2005}. 

However, there are also structural barriers to healthcare utilization and related inequitable treatment that disproportionately affect women and minorities \cite{schulman_Effect_1999, chapman_Physicians_2013, lee_RacialEthnicDisparities_2019}. For instance, Black children in the United States receive fewer pain management medications after surgery compared to White children \cite{sabin_Influence_2012}, and women receive worse treatment for coronary heart disease than men \cite{healy_Yentl_1991, schulman_Effect_1999}

Patient data samples tend to disproportionately represent wealthier population segments and under-represent those from lower socio-economic backgrounds across societies \cite{oecd_Health_2019}. Health disparities follow social class indicators such as education, income, occupation, and wealth are strongly linked to health outcomes across nations. Taken together, these factors create a paradoxical health data poverty in the age of big data \cite{ibrahim_Health_2021}. The data representativeness problem poses a risk for AI automation that could increase the existing health divide \cite{obermeyer_Dissecting_2019, amsterdam_When_2025}.

Further insights into patient data generation come from an examination of clinicians as controllers of patient care and patient data collection. In control systems theory, controller actions are understood to be guided by process models \cite{leveson_Engineering_2016}. This means that their actions are informed by their goals, mental models of the process, knowledge of actions they can enact, and the feedback which they can receive from those actions. 

Patient care is the primary task performed by clinicians, and data generation is a secondary task shaped by reporting, communicating, and billing needs \cite{schulz_Standards_2019, europeanunion_Regulation_2025}. Both tasks can be linked to control systems theory: Patient care decision-making is linked to clinical reasoning \cite{higgs_Clinical_2019}, and can be understood through clinician mental models of the care process. In clinical care, patient data are collected to ascertain patient health states. This is related to the control theory notion of feedback (sometimes called the \textit{observability condition}; \cite{leveson_Engineering_2016}). 

Patient data collection accounts for a part of patient data generation that is within clinician control (as opposed to mechanisms that give rise to disease). It can be understood as a complex sociotechnical system. Some of its aspects are rife with subjective decisions \cite{olson_What_2021}, including which aspects to investigate and which to dismiss, how to interpret signs, symptoms and measurements \cite{scott_errors_2009, schiff_Diagnostic_2009, norman_Diagnostic_2010}, and what to note down in the patient record \cite{pakhomov_Agreement_2008}. Other aspects of patient data generation are shaped by the characteristics of upstream medical devices, instruments, information storage systems, and software. All of these factors leave marks on downstream patient datasets.

Patient data collection occurs through \textit{observation} \cite{faustinella_Power_2020}, \textit{report} \cite{pakhomov_Agreement_2008}, and \textit{measurement} \cite{vet_Measurement_2011}. Patient signs are visible. They can be easily observed and reported by healthcare professionals, care-givers, or patients themselves. Patient symptoms, however, are invisible. They can only be reported by patients. Clinical measurements enable gathering of additional information with the help of technology. They quantify some otherwise latent aspects of patient health \cite{wilson_Linking_1995}. 

Patient information artifacts collected in these ways often span across modalities. They are typically stored in digital information systems, such as computerized physician order entry systems or electronic health records. In AI automation contexts, these artifacts are considered raw data that might become data-analytic model inputs. The term 'real-world data' is sometimes used in this context. It highlights cases where data are produced by naturally occurring processes (i.e. patient care) rather than through experimental design or simulation.

Characteristics of patient care profoundly shape data which it generates. For instance, consider that patient care often spans multiple settings, with different clinicians and institutions involved at various stages of clinical care. As a consequence, clinical decision-making and patient data generation tend to be highly compartmentalized \cite{kern_Care_2024}. The fragmented nature of patient care often leads to data gaps where any given patient care setting may only have a small fraction of patient data following a health event. In some cases, feedback is missing and this limits control exerted by clinicians and AI systems.

As an example, consider the workflow of a clinical neurophysiologist who interprets EEG measurements. The interpretation process typically results in a conclusion letter (for instance expressing the degree to which epileptic abnormalities were detected). In fragmented patient care settings, this letter is subsequently forwarded to the patient's referring neurologist, possibly at another institution. In this case, the neurologist is tasked with making diagnostic and treatment decisions. It is not guaranteed that either of these decisions will be communicated back to the clinical neurophysiologist nor encoded in their patient records. 

Similarly, fragmentation of patient care means that patient outcomes that occur further downstream (e.g. recovery, worsening, or side-effects) may be completely missing from patient data. This results in a situation where clinicians may lack feedback on their decisions (e.g. whether a 'negative decision' was a 'true negative' or a 'false negative'). Similarly, data-analytic models trained on data that lack downstream patient outcomes cannot outperform current human clinicians when labels made by the same clinicians represent the only ground truth outcome label that is available.

Another key factor that we need to consider is that patient care practices vary greatly both within and between patient care settings, leading to significant heterogeneity of patient data. This outcome is caused by both human and technological factors operating at any patient care setting. Data are simply captured in different ways which can pose a threat to generalizability of data-analytic models in patient care, assuming that such AI system transferability is even desirable \cite{selbst_Fairness_2019}. Data heterogeneity can also reduce statistical power of data-analytic models.

As an example, consider the common case of multiple clinicians who work at the same patient care department and share control over patient data generation. Analyses of real-world patient data have shown that clinicians differ widely in their use of terminology to describe same underlying medical conditions and procedures \cite{bitterman_Clinical_2021, coghlan_Danger_2023}. Data instances may appear distinct in spite of capturing the same underlying conditions. Data-analytic models trained on heterogeneous data risk having low statistical power to detect the target condition. An easy way to overcome this issue is through feature engineering that incorporates domain knowledge.

\begin{figure}[h]
\centering
\caption{Most real-world patient datasets arise from a mixture of distinct data-generating processes. Linking individual data instances to those clinical processes which had generated them is not always straightforward. Indicators of clinical processes may be implicit, latent, and not clearly understood by all stakeholders which engage with patient data.}
\label{fig:mixture_dgps}
\begin{center}
\includegraphics[width=0.6\textwidth]{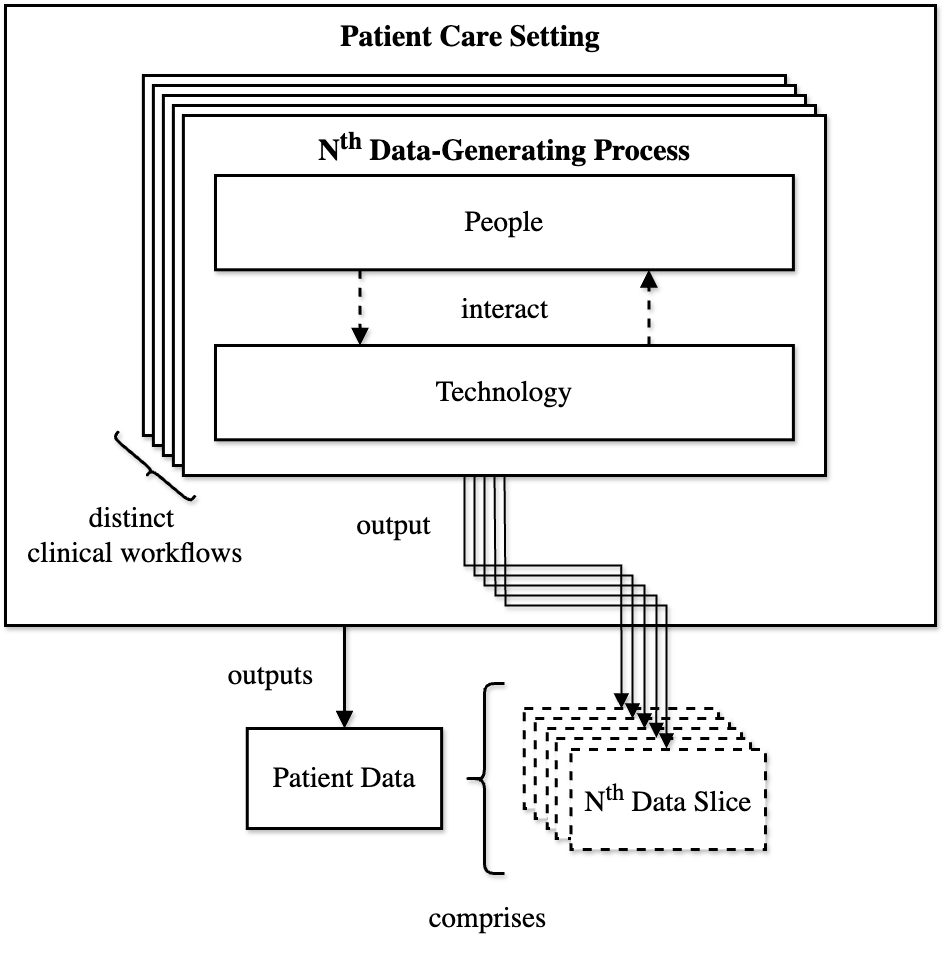}
\end{center}
\end{figure}

Patient data reflect operations of the environments which produce them. Most patient care settings comprise multiple workflows, each associated with a distinct patient population that is explained through a different data generating mechanism (Figure \ref{fig:mixture_dgps}). 

For example, consider a patient dataset containing electroencephalogram (EEG) measurements originating from a hospital neurology department. A patient could appear in the associated EEG measurements dataset if they had followed the diagnostic process after experiencing a first seizure-like event. This could be thought of as the epilepsy diagnosis workflow. However, other patients could have followed completely different clinical processes such as the diagnostics of cognitive disorders, dementia, encephalopathy, or delirium. The EEG dataset may contain all these instances, but these instances need not share their data generating mechanism.

The very nature of clinical measurements and their interpretation varies by the intended purpose (i.e. clinical question) that had guided the data-generating process. Patients associated with different workflows likely differ along a range of key characteristics. For example, patients investigated for dementia are likely to be systematically older than patients investigated for epilepsy after a first seizure-like event. When instances in a dataset come from a mixture distribution (Figure \ref{fig:mixture_dgps}), using all instances to train a model could render its conclusions invalid for specific applications. 

Data coming from a mixture of distributions is an example of a patient data characteristic that is difficult to discern through data analysis alone. While clustering approaches such as latent class analysis can explore the multivariate feature space for similarities among the instances, the validity of such approaches is still greatly improved by the incorporation of domain knowledge \cite{vanlissa_Recommended_2024}. 

However, patient datasets often lack clear documentation and metadata reflecting the workflows which had produced the data, and the contexts in which they are most applicable \cite{pushkarna_Data_2022, hupont_Use_2024}. Such information is sometimes latent or unclear especially to data scientists with limited clinical expertise. In many cases, it is only discovered through a close collaboration with clinical domain experts. The lack of awareness about distinct clinical processes which are captured in the data is a significant threat to validity of AI automation.

\begin{figure}[h]
    \begin{center}
        \caption{In clinical decision-support, clinicians and AI systems share control over decision-making. A feedback loop can occur when prior patient decisions shaped by AI systems are used as inputs to inform future automated decision proposals.}
        \includegraphics[width=0.53\textwidth]{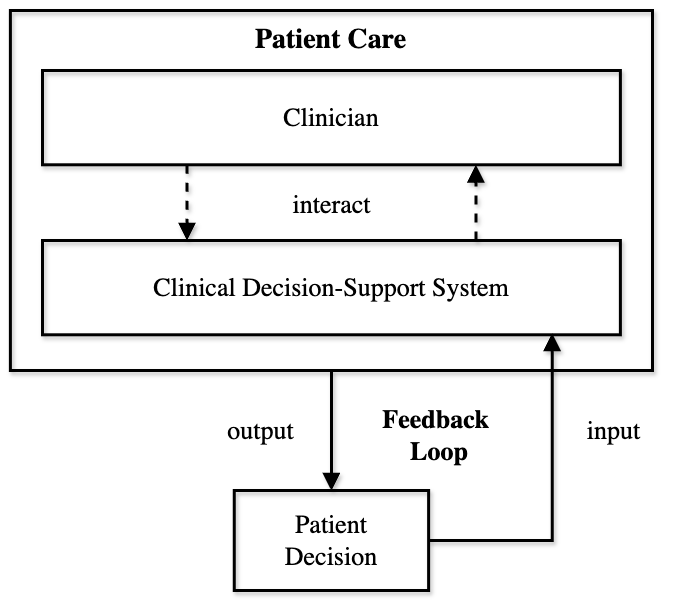}
        \label{fig:feedback}
    \end{center}
    \end{figure}

Finally, we would like to reflect on how the deployment of AI systems into patient care may shape the downstream patient data generation. In clinical decision-support, data-analytic model outputs (e.g. treatment recommendations) are influenced by their inputs (i.e. historical patient data), but final decisions are made by clinicians.

Here, previous patient treatment decisions influence present model outputs and model suggestions exert some influence on clinician decision-making and thus on the downstream patient outcomes \cite{green_Disparate_2019}. 

In control systems theory, the dependence of the current model outputs on the previous data is an example of a \textit{feedback loop} (Figure \ref{fig:feedback}). Feedback loops can be problematic when they lead to an amplification of undesirable data properties: Small initial disparities can become greater over time through the use of AI systems \cite{taori_Data_2023}. 

In deployment, feedback loops could lead to the propagation of patient harms \cite{amsterdam_When_2025}. They can occur even with human oversight, i.e. when AI systems are not granted full autonomy. From a data-generation perspective, AI automation creates a dependency of newly created patient data on the properties of the past data.

Taken at face value, data represent some parts of reality \cite{barocas_Fairness_2023}. However, it is good to wonder whether some of its aspects might not be worth mirroring through model representations. There are certainly data biases that can cause allocative and representational harms. However, the challenge is that while all biases are differences, not all differences are biases. In complex landscapes such as those found in healthcare, differences may also arise due to biological mechanisms that cause conditions, explain their prevalence, presentation, and responsiveness to treatment. 

In such cases, data analysis alone is often insufficient to tell apart the signal form the noise \cite{anadria_Algorithmic_2024}. Responsible AI system development has to turn to domain knowledge \cite{sogancioglu_Fairness_2024} and purpose-driven design to explore how to translate real-world problems into trustworthy model representations.

\section{Data We Want: Systems Theory for Data Quality Requirements}
\label{sec:data-we-want}

\subsection{Context}
\label{sec:data-we-want-premise}
In this section, we turn our focus away from patient data generation, and onto the AI automation task. We argue that systems thinking can help uncover data quality requirements based on the intended purpose of AI automation, and thus support a purpose-driven data-analytic model development.

Our argument is built on two distinct yet complementing streams of systems research. In particular, we adapt ideas from the \textit{system-theoretic process analysis} \cite{leveson_Engineering_2016, leveson_STPA_2018}, a method that emerged in control engineering and system safety discipline, and the \textit{conceptual framework for systems analysis in multi-actor policy research} \cite{enserink_Policy_2022}. While emerging in different contexts, both of these methods rely on the principles of systems thinking. To our knowledge, neither approach has previously been applied to the study of data-analytic model development. We complement these system-analytic approaches with ideas stemming from research design in social sciences. This methodology helps meet our analytic aim.

Our aim is '\textit{to explain how AI automation objectives shape data quality requirements}'. We approach this as a theoretical exercise where the goal is to understand the logic and the structure of this problem for any data-analytic automation task in patient care. 

To meet this aim, we define a general intended purpose of AI automation to serve as a surrogate for the more refined forms this definition should find in applied cases. For our analytic aim, it suffices to define the intended purpose of AI automation as: '\textit{to transform manual patient care delivery through the introduction of data-analytic models into an operational environment}'. 

This definition is purposefully open-ended. In applied contexts, its formulation will acquire much more detail. Some of the likely addition to this purpose are 'to improve or at least maintain health outcomes' and 'to maintain or reduce the required resources'. Consider that factors such as the nature of the clinical task \cite{higgs_Clinical_2019}, the desired level of automation \cite{vagia_Literature_2016}, the available means, constrains, and values of those involved in system design negotiations \cite{enserink_Policy_2022} all influence AI system requirements and constraints.

In fact, the formulation of the intended purpose for an AI system may vary by stakeholder \cite{enserink_Policy_2022}. While this plurality of perspectives adds complexity to AI system development, it also brings immense value to its design. Consider that each individual perspective may capture a small part of the greater intended purpose. Therefore, each relevant stakeholder may help refine the formulation of AI system requirements and constraints, thus aiding its design. 

As alluded to in Sections \ref{primer-systems-theory} and \ref{sec:premises}, the key premise of this section is that AI automation should be understood as a complex engineered system that augments the delivery of patient care through data-analytic models and patient data. This enables the application of systems theory to elucidate how the interaction of various automation system components gives rise to its outcomes \cite{leveson_Engineering_2016}. We initially follow the standard steps of systems analysis, to draw the system boundaries and define its structure \cite{leveson_STPA_2018}. However, instead of focusing on the entire system, our goal is to focus on its data component. 

\subsection{A Systems Perspective on AI Automation}
\label{sec:data-we-want-theory}

\begin{figure}[h]
\begin{center}
        \caption{A system-theoretic view of AI automation process components and quality evaluation procedures.}
        \includegraphics[width=0.75\textwidth]{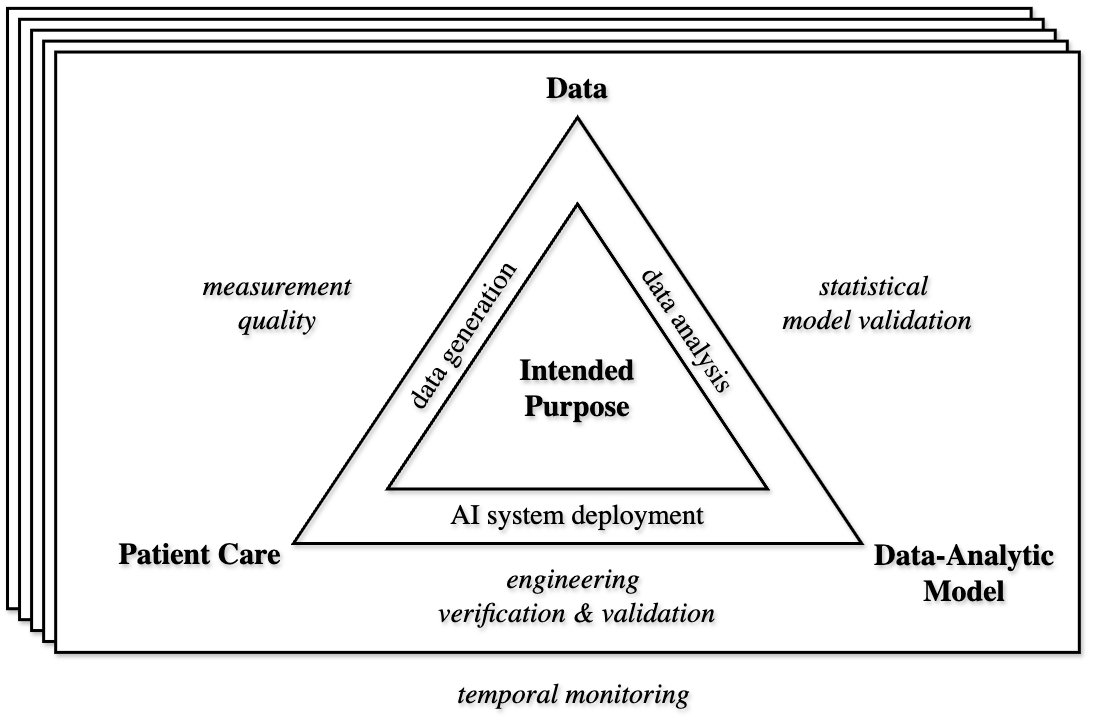}
        \label{fig:triangle}
\end{center}
    \end{figure}

Our systems analysis of data requirements for AI automation follows several steps. Its goal is to explain the functional and quality requirements of data as AI automation system components. 

The first step in a systems analysis is to draw the system boundaries which define what components are being considered in the analysis. Based on our analytic aims, we include: \textit{patient care}, \textit{data}, and the \textit{data-analytic model} components.

The following step in systems analysis is to map the structure of the modeled system \cite{leveson_STPA_2018}. This refers to capturing the functional relationships and interactions of the modeled system components \cite{leveson_STPA_2018}. 

Since our AI automation system has three components, we map them in a triangular configuration (Figure \ref{fig:triangle}). In the center of the triangle, we place 'intended purpose' as the reminder that a system cannot be defined, studied, nor evaluated without a clear objective \cite{leveson_STPA_2018}. It acts as a reminder that purpose-driven AI system design has to meet requirements of many stakeholders. For this theoretical exercise, we hold on to the general formulation of the intended purpose and ask: 'What functions does this component perform relative to the other two components?'. 

This question starts revealing the structure of the AI automation system. We know that patient care collects patient data and that patient data measure aspects of patient care. This process corresponds to patient data generation which we discussed in Section \ref{sec:data-we-have}. 

Next, we know that patient data interface data-analytic models in two ways. The first way occurs in model development where patient data are used as inputs for the training or fine-tuning of data-analytic models. The second way occurs in model operation where patient data are simply processed using a deployed model to derive its outputs. We refer to both of these cases data analysis (revisit Figure \ref{fig:statistical-inference}).

Finally, data-analytic models are applied to ongoing patient care. On their own, models serve as theories about clinical decision-making, but in deployment, they act as controllers of its processes. We refer to this process as AI system deployment. Following this line of reasoning, we have uncovered the structure of AI automation. 

Systems analysis of the structure of AI automation has revealed that this complex engineering problem consists of three inferential steps connecting patient care, patient data, and data-analytic models: data generation, data analysis, and AI system deployment.

Since our analytic aim is to better understand the data component, we further focus on its function as a part of the AI automation system (Figure \ref{fig:triangle}). Data can be seen as measurements of patient care which serve as a representation of clinical decision-making that is learned in model development and applied in AI system operations. Since specifies what patient data should do as AI automation system components, it can be understood as stating their functional requirements \cite{enserink_Policy_2022}. 

In systems engineering, functional requirements are supported by quality requirements which specify how well the component should perform its function. Based on the general-formulation of the intended purpose of AI automation, we can move from understanding the function which data perform within the system to specifying their quality requirements:

\begin{quote}
\begin{center}
\texttt{Patient data have to measure the right components of patient care, as defined by the AI automation objectives, and to do so sufficiently well to support data-analytic modeling in reaching clinically legitimate statistical inference.}
\end{center}
\end{quote}

The system-theoretic formulation of data quality requirements for AI automation is closely aligned with that of the EU AI Act (Article 10) \cite{europeancommission_Regulation_2024}. The notion of 'measuring the right components of patient care' closely corresponds to data quality dimensions of relevance and representativeness. The notion of doing so 'sufficiently well' corresponds to the dimensions of 'free of error' and 'complete'. The 'intended purpose' of the data component can be understood as 'to support data-analytic modeling in reaching clinically legitimate statistical inference'. 

We observe that the desired data qualities for AI automation closely match measurement theory constructs of \textit{validity} ('to measure the right components') and \textit{reliability} ('to do so sufficiently well') \cite{vet_Measurement_2011}. \textit{Measurement theory} is a fundamental branch of scientific theory which concerns itself with the nature and quality of measurement. It developed in research design where theory-building and construct operationalization are given a lot of attention. 

Measurement theory wrestles with some of the key epistemological problems that arise when modeling reality. Since many natural phenomena are not directly observable, it focuses on proposing ways how such latent constructs can be represented legitimately and provides a language needed to express different aspects of measurement quality (Figure \ref{fig:triangle}). 

For instance, if data are viewed as a collection of measurements, then they have \textit{content validity} if their scope captures all the relevant facets of the modeled construct \cite{vet_Measurement_2011}. In patient care, this could mean having all the relevant features for the target task available in the dataset, which can be understood as completeness for the intended purpose that is focused on data dimensions rather than on instances. 

Another measurement quality is \textit{criterion validity} which data have if they relate to or predict the target outcome. \textit{Cross-cultural validity} evaluates whether data that were generated in one setting are meaningful for another. This clearly links to concerns of model generalizability or portability \cite{selbst_Fairness_2019}.

In spite of its centrality, measurement quality is rarely considered in model development. Computer scientists and responsible AI researchers are seldom aware of measurement theory \cite{jacobs_Measurement_2021}. This lack of awareness represents a broader culture shift that took place in data analysis over the last two decades. 

In his seminal paper which we quoted earlier, Breiman \cite{breiman_Statistical_2001} described the emergence of data-driven modeling and praised its potential to revolutionize data analysis. Since then, data-driven approaches, including machine learning and deep neural networks, have only grown in popularity and become a staple of AI automation of patient care. 

In this application domain, and in others, this reliance on data quantity came at the expense to considerations of data quality. For cultural rather than technical reasons, the value of domain theory and design principles has reduced in many computational branches of data analysis. However, we have argued that computational approaches alone are often insufficient to ensure clinical legitimacy of model representations. Same applies for algorithmic fairness methods: When model developers lack an understanding of the real-world, bias mitigation risks "throwing the baby out with the bathwater" \cite{anadria_Algorithmic_2024}. 

Machine learning models and deep neural networks can certainly benefit from the incorporation of measurement quality constructs in their development \cite{jacobs_Measurement_2021,du_Assessing_2021,fang_Designing_2023,fang_Leveraging_2024}. What we see as the key barrier for the wider adoption of clinical measurement theory in AI automation is that this requires a deep understanding of the subject matter that is being modeled. In other words, it requires system thinking and purpose-driven design.

 In patient care, this means both the knowledge of the data generation process and the clinical domain. We should study process models and clinical decision-making in ways that can inform data requirements and support AI automation. Current model evaluation practices largely emphasize a model's predictive performance on the held-out validation set \cite{breiman_Statistical_2001}. This goodness-of-fit measure evaluates how well the model fits the data. While statistical model validation (Figure \ref{fig:triangle}) is valuable for the evaluation of statistical inference, the sole focus on this inferential step fails to answer: How well does the model fit reality?

The systems perspective therefore makes it evident that the field of AI has de-emphasized the value of purposeful design, and put data quantity above data quality. Ever since the dawn of data-driven modeling, the argument has been that one need not concern themselves too much with the learned model representation -- it would simply emerge from the data \cite{breiman_Statistical_2001}. However, understanding the real-world data generation process means knowing just how noisy and biased data can be and how limited it is what they measure (Section \ref{sec:data-we-have}). The data quantity-over-quality bias in contemporary machine learning has is limiting to both the validity of statistical inference and to the objectives which AI automation systems can meet. 

Our key finding is that AI automation requires systems thinking. Instead of just `throwing more data at models', both data and model pipelines need to be purposefully designed. This design, in turn, has to be informed by theory. 

Systems thinking makes it clear that AI automation requires theory about that which it automates. To measure the right components of patient care means to capture those components of reality that align with the intended purpose of automation. If the intention is to predict a medical condition, then the data should contain predictors of this medical condition. If the intention further specifies that the AI system should perform early detection, then the available predictors should represent those early predictive signs that are available or possible to collect on time for this objective. 

AI systems need purposeful design, and this design can be supported by systems thinking and clinical measurement theory. Fundamentally, since AI systems act as controllers, the system design should start by asking: \textit{What is known about this patient care task which we want to automate?} 

\section{Discussion}
\label{sec:discussion}

AI automation could be a solution to tackle many challenges facing patient care \cite{michaeli_Healthcare_2024}. Machine learning could reduce clinical workloads and operating costs \cite{patel_ChatGPT_2023, agaronnik_Challenges_2020}, all the while increasing provider productivity \cite{kochling_Discriminated_2020} and delivering state-of-the-art predictive and preventative care. This potential can be maximized through purpose-driven design.

\subsection{What should machine learning model development look like?}

We posit that machine learning model development for automation of patient care should start from the intended purpose of the AI system that is being developed. This purpose-driven way of reasoning is in line with systems engineering and design approaches \cite{leveson_Engineering_2016, leveson_STPA_2018}, but has become de-emphasized in parts of contemporary machine learning. Taking real-world data at face value and exploring them for automation utility may lead to functioning AI systems. However, this approach is also inherently limiting our imagination of what is possible to achieve with machine learning.

Purpose-driven model development should start with the intended purpose of the AI system. This means that model developers ought to have a strong understanding of the stakeholder expectations and realities of operational settings undergoing automation. Since individual stakeholders may see different elements of the greater intended purpose, participatory design approaches could enable the capturing of all the relevant AI system requirements and constraints \cite{enserink_Policy_2022}. 

Some controls could be enacted through institutional mechanisms external to the machine learning system \cite{dobbe_System_2024}. Other requirements and constraints may affect the design of machine learning systems and their data pipelines. These are especially important for model developers. As an example of a requirement affecting model design decisions, consider that if a model is to perform early detection of a clinical condition, its development should imitate the temporal revelation of new patient data, that is simulate the placement of the model in ongoing patient care workflows.

While requirements engineering may be necessary for responsible automation of patient care, its reintroduction should not replace the existing machine learning methodologies. Much of the existing functionality of AI systems can be attributed to the major innovations that data-driven approaches have brought to data analysis. We highlight the value of deep neural network architectures and other non-parametric model specification, predictive inference on a held-out validation set, loss optimization techniques and algorithmic fairness methods. Machine learning should keep these valuable elements and complement them with the valuable ideas from the theory-driven culture. We especially highlight the focus that this tradition has given to validity of its inputs through reliance on clinical theory, data collection and control of measurement error.

\subsection{How to understand patient data quality?}

Much of our focus was on understanding the role of patient data for AI automation of patient care. If one were to compare the \textit{data we have} to the \textit{data we want} for any particular patient dataset, they would be performing a purpose-driven data quality evaluation. In our view, this is the only meaningful way to understand patient data quality for AI automation.

While data quality can be expressed through various facets \cite{batini_Methodologies_2009, bicevskis_Step_2019, bronselaer_Data_2021, mohammed_Data_2024}, we argue that the aggregate is not best understood as a sum of its parts. Patient data quality is a complex emergent property that is shaped by both the convoluted data-generating mechanisms of patient care, and the complex demands that its automation dictates. This means that patient data quality assessments have to also be approached through an understanding of data generation and automation requirements.

The European Union (EU) AI Act \cite{europeancommission_Regulation_2024} discusses data quality requirements for high-risk AI systems in Article 10. These requirements are directly applicable to patient data used for AI systems performing clinical decision-support \cite{vankolfschooten_EU_2024}. The Act emphasizes that data quality should be understood in relation to the intended purpose of the AI system, that is the use that was intended by the system provider. In Article 10, data quality is discussed through various dimensions including model and data pipeline design, measurement qualities, data biases, etc. System-theoretic approaches could help form the harmonized standard for this task, but specific methods remain to be proposed.

\subsection{How should AI automation shape the design of patient data pipelines?}

Patient care providers have significant control over their data quality, especially where they control patient data collection. From a systems lens, they can improve data quality through interventions to data collection and data processing. Some of the most impactful interventions could improve standardization, interoperability, labeling of clinical workflows and metadata, and closing the existing data gaps, especially as related to patient outcomes and social determinants of health. The knowledge of tasks which providers want to automate can help plan the redesign of data pipelines that can support these objectives. Here, more data is not always better data.

Policy makers can also support data quality efforts for AI automation of patient care. The recent European Health Data Space initiative seeks to foster interoperability of patient records across the European single market as part of the broader European strategy for data \cite{europeanunion_Regulation_2025}. While current patient care \cite{schulz_Standards_2019} and regulatory perspectives \cite{europeanunion_Regulation_2025} see AI automation as a secondary use of patient data, we expect a future where AI automation plays an increasing role in the delivery of patient care. This future requires a purposeful design of patient data pipelines, cognizant of various automation objectives.

\needspace{5\baselineskip}
\subsection{How could future work contribute to purpose-driven AI automation?}

This work is only scratching the surface of how machine learning for AI automation of patient care may benefit from systems thinking. By establishing bridges between data analysis and systems theory, we have attempted to explain patient data quality as a function of data generation and AI automation objectives. This approach brings valuable insights for machine learning model development. However, AI automation is highly complex and this work has addressed patient data which are a rather small component of the greater picture. 

Our choice to focus on the data component is informed by the current clinical reality: Many clinical care providers possess large quantities of patient data, and wish to automate their processes, yet lack clarity about the utility of these data for AI systems. We found a combination of systems thinking and data analytic theory as a helpful frame to bring understanding to some of the key challenges associated with patient data and AI automation of patient care. This appears to be a promising direction for future work which could:

\begin{itemize}
    \item Expand the theoretical framework beyond data onto other components of patient care.
    \item Propose methods which model developers can use to understand the quality of their current data and the requirements set by the automation task.
    \item Explore how well this approach generalizes to sectors other than patient care.
    \item Test the utility of these ideas onto concrete AI automation problems.
\end{itemize}

\section{Conclusion}
\label{sec:conclusions}

In this work, we have reflected on two data-centric challenges affecting AI automation of patient care. We have studied patient data by both looking upstream towards their generation, and downstream towards their purpose for AI automation. This required a novel combination of systems thinking and data-analytic theory. 

We argue that machine learning in automation of patient care can benefit from the incorporation of the sociotechnical context surrounding data-generation and AI system deployment. Systems thinking can complement the existing machine learning methods and help understand data requirements. For AI automation of patient care, patient data should measure the right components of patient care, and that they do so sufficiently well, to support data-analytic modeling in reaching clinically legitimate statistical inference. 

While this does not invalidate the existing predictive performance and algorithmic fairness approaches in machine learning, but it calls for greater attention to model inputs during model development. We highlight theory-driven modeling as a source of inspiration for the role that measurement theory and systems thinking can play for responsible model development. Taken together, our findings point to the need for purpose-driven design where machine learning is shaped by systems thinking and clinical measurement theory. 

Furthermore, we have argued that patient care stakeholders may vary in their formulations of AI system requirements and constraints, and that their diverse perspectives ought to shape machine learning model development. Our contribution represents a small step towards a new perspective on AI automation of patient care. Many of the points we raise may generalize to other safety-critical domains. The purpose-driven perspective opens up many opportunities and challenges for future work which are linked both to automation theory and model development methods. We hope that future work continues this line of inquiry to pave the way for safe automation of patient care.

\section{Acknowledgments}
In addition to those formally recognized as co-authors, the first author wishes to thank Dr. Luuk Wieske and Dr. ir. Vera Lagerburg of St. Antonius Hospital in Utrecht, Dr. Seda Gürses (Programmable Infrastructures Group at Delft University of Technology), Dr. Fanny Jourdan (IRT Saint Exupéry), Dr. Stephanie 'Ace' Medlock (Amsterdam University Medical Center), Dr. Hannah van Kolfschooten (University of Amsterdam), Martijn Logtenberg (Utrecht University), Jacqueline Kernahan and other members of the Sociotechnical AI Systems Lab at Delft University of Technology. Your inputs have all offered direction. Also to my statistics teachers, Prof. Dr. Ellen Hamaker, Dr. Noémi Schuurman, Dr. Gerko Vink, and Dr. Erik-Jan van Kesteren of Utrecht University. Your thought has greatly shaped my understanding of the world, and I am grateful to have learned from you. Thank you. This work is because you are.

This work was supported by the Dutch Research Council (Dutch: Nederlandse Organisatie voor Wetenschappelijk Onderzoek; NWO) under project number 024.005.017. The views expressed and arguments employed herein are solely those of the author(s) and do not necessarily reflect the views of the OECD or its member countries.

\needspace{5\baselineskip}
\bibliographystyle{unsrt}  
\bibliography{references}


 \end{document}